\documentclass[sigplan,screen]{acmart}
\AtBeginDocument{%
  \providecommand\BibTeX{{%
    \normalfont B\kern-0.5em{\scshape i\kern-0.25em b}\kern-0.8em\TeX}}}

\setcopyright{acmcopyright}
\copyrightyear{2023}
\acmYear{2023}
\acmConference[KDF @SIGIR '23]{The 4th Workshop on Knowledge Discovery from Unstructured Data in Financial Services}{July 27,
  2023}{Taipei, Taiwan}
\begin{document}

\title{GPT-FinRE: In-context Learning for Financial Relation Extraction using Large Language Models}

\author{Pawan Kumar Rajpoot}
\authornote{Both authors contributed equally to this research.}
\email{pawan.rajpoot2411@gmail.com}
% \author{Ankur Parikh}
\authornotemark[1]
\affiliation{%
  \institution{MUST Research}
  \streetaddress{P.O. Box 1212}
  \city{Bangalore}
  \state{Karnataka}
  \country{India}
}

\author{Ankur Parikh}
\email{ankur.parikh85@gmail.com}
\affiliation{%
  \institution{UtilizeAI Research}
  % \streetaddress{1 Th{\o}rv{\"a}ld Circle}
  \city{Bangalore}
  \state{Karnataka}
  \country{India}}

\begin{abstract}
Relation extraction (RE) is a crucial task in natural language processing (NLP) that aims to identify and classify relationships between entities mentioned in text. In the financial domain, relation extraction plays a vital role in extracting valuable information from financial documents, such as news articles, earnings reports, and company filings.
This paper describes our solution to relation extraction on one such dataset REFinD. The dataset was released along with shared task as a part of the Fourth Workshop on Knowledge Discovery from Unstructured Data in Financial Services, co-located with SIGIR 2023. In this paper, we employed OpenAI models under the framework of in-context learning (ICL). We utilized two retrieval strategies to find top K relevant in-context learning demonstrations / examples from training data for a given test example. The first retrieval mechanism, we employed, is a learning-free dense retriever and the other system is a learning-based retriever. We were able to achieve 3rd rank overall (model performance and report). Our best F1-score is 0.718.

\end{abstract}




\maketitle

\section{Introduction}
\begin{figure}
\includegraphics[width=\linewidth,scale=2,]{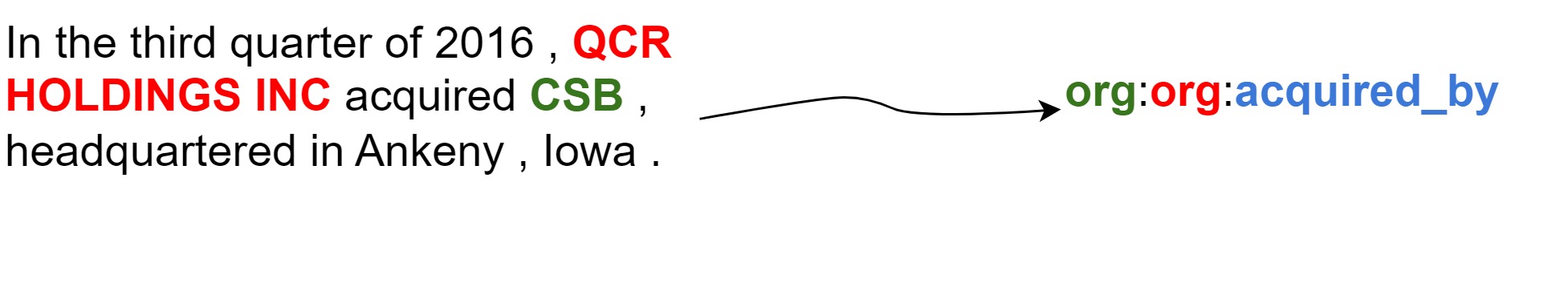}
\caption{Relation Extraction example, here both organizations are connected with "acquired by" relation.}
\label{fig:Entity and Relation types in REFinD dataset}
\end{figure}
The emergence of large language models (LLMs) such as GPT-3 \cite{brown2020language}\cite{thoppilan2022lamda} represents a significant advancement in natural language processing (NLP). These models have expertise in variety of domains and hence they can be used as it is in multiple NLP tasks.
Traditionaly language models use separate pre-training-and fine-tuning pipelines \cite{devlin-etal-2019-bert} \cite{beltagy-etal-2019-scibert} \cite{raffel2020exploring} \cite{lan2020albert} \cite{zhuang-etal-2021-robustly} where fine-tune stage follows pre-training. Models are fine-tuned on a task-specific dataset in a fully-supervised manner.
More recently a new paradigm known as in-context learning (ICL) \cite{brown2020language}\cite{min2022rethinking} is being used which formulates an NLP task such that LLMs make predictions by learning from demonstrations. These demonstrations are presented to the LLMs in the context prompt itself. 

  Under the framework of ICL, LLMs achieve remarkable performance rivaling previous fully-supervised methods even with only a limited number of demonstrations provided in the prompt in various tasks such as solving math problems, commonsense reasoning, text classification, fact retrieval, natural language inference, and semantic parsing \cite{brown2020language} \cite{min2022rethinking} \cite{DBLP:journals/corr/abs-2104-08768}. Recently, ICL based approach\cite{wan2023gpt} is utilized for Relation Extraction (RE) task. RE seeks to identify a semantic relationship between a given entity pair mentioned in a sentence, which is the central task for knowledge retrieval requiring a deep understanding of natural language. The approach achieves improvements over not only existing GPT-3 baselines, but also on fully-supervised baselines. Specifically, it achieves SOTA performances on the Semeval and SciERC datasets, and competitive performances on the TACRED and ACE05 datasets.
  
  Retrieval of examples to demonstrate is a key factor in the overall performance on these pipelines. LLMs can relate to the presented "to be predicted" data point more if the contextual examples predicted are similar to it. More relevant examples help us to leverage more out from LLMs both in terms of improvement in performance and less hallucination as examples can demonstrate model not to hallucinate in some cases.   
  
  In this paper, we employed GPT-3.5 Turbo and GPT-4 under the framework of ICL for the relation extraction task on REFinD dataset. We utilized two retrieval strategies to find top K relevant in-context learning demonstrations / examples from training data for a given test example. The first mechanism we have employed is a learning-free dense retriever. The other system we have utilized is a learning-based retriever \cite{rubin-etal-2022-learning}.   
\section{Preliminary Background}

\subsection{ Task Definition}
As per the challenge "Relation Extraction is the task of automatically identifying and classifying the semantic relationships that exist between different entities in a given text."
This shared task is a part of "Knowledge Discovery from Unstructured Data in Financial Services" (KDF) workshop which is collocated with SIGIR 2023.

Let C denote the input context and e1 in C, e2 in C denote the pair of entity pairs. Given a set of predefined relation classes R, relation extraction aims to predict the relation y in R between the pair of entities (e1, e2) within the context C, or if there is no predefined relation between them, predict y="no relation".

\subsection{Data}

The dataset \cite{kaur2023refind} released with this task is the largest relation
extraction dataset for financial documents to date. Overall REFinD
contains around 29K instances and 22 relations among 8 types of entity pairs.
REFinD is created using raw text from various 10-X reports (including
10-K, 10-Q, etc. broadly known as 10-X) of publicly traded companies obtained from US Securities and Exchange Commission. Figure-2 shows different entity types and relations exist between them.

\begin{figure}
  \includegraphics[width=\linewidth]{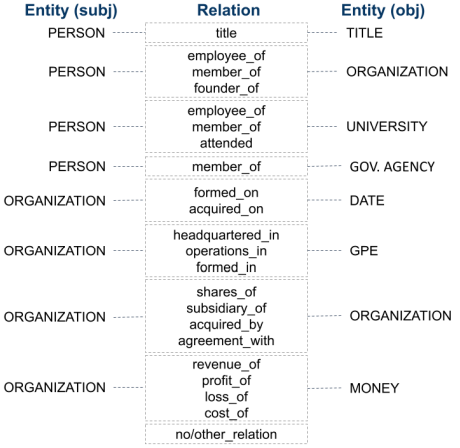}
  \caption{REFinD dataset relation and entity types.}
  \label{fig:Entity and Relation types in REFinD dataset}
\end{figure}

\subsection{In Context Learning}
In-context learning (ICL) refers to one of the core emergent abilities \cite{wei2023larger} that infers new tasks from context. We use the terms 'in-weights learning' and 'in-context learning' from prior work on sequence models \cite{brown2020language} to distinguish between gradient-based learning with parameter updates and gradient-free learning from context, respectively. Formally, each training instance is first linearized into an input text x = (x1...xn ) and an output text y = (y1...yn), where for all tokens x1...xn, y1...yn in V and V is the vocabulary set of the LM. Given a new test input text x-test, in-context learning defines the generation of output y as y-test $\sim$ PLM(y-test | x1,y1,...,xk,yk, x-test), where $\sim$  refers to decoding strategies (e.g., greedy decoding and nuclear sampling \cite{li2022contrastive}), and each in-context example ei= (xi,yi) is sampled from a training set D. The generation procedure is especially attractive as it eliminates the need for updating the parameters of the language model when encountering a new task, which is often expensive and impractical. Notably, the performance of ICL on downstream tasks can vary from almost random to comparable with state-of-the-art systems, depending on the quality of the retrieved in-context examples \cite{rubin-etal-2022-learning} \cite{DBLP:journals/corr/abs-2101-06804} \cite{wu2023selfadaptive}. 

\begin{table*}[ht]
\centering
% \begin{tabular}{lcr}
\begin{tabular}{|*{3}{p{\dimexpr0.33\linewidth-2\tabcolsep-1.2\arrayrulewidth\relax}|}}

\hline
\textbf{Retriever} & \textbf{LLM} & \textbf{F1-Score}\\
\hline
\verb|KNN with openAI embeddings| & {GPT 3.5 Turbo (Examples: 5 retrieved + 5 random per possible relation)} & {0.643}\\
\verb|KNN with openAI embeddings| & {GPT 4 (Examples: 5 retrieved + 5 random per possible relation)} & {0.697}\\
\verb|EPR with GPT-Neo-2.7B| & {GPT 4 (Examples: 2 retrieved + 3 random per possible relation)} & {0.703} \\
\verb|EPR with GPT-Neo-2.7B| & {GPT 4 (Examples: 5 retrieved + 4 random per possible relation)} & {0.718} \\
\hline
\end{tabular}
\caption{Our performance on test data with different combinations of retriever and LLM}
\label{tab:test_metrics}
\end{table*}

\section{GPT-FinRE}
GPT-RE is formalized under the ICL framework, using GPT models as shown in Figure-3. 
% \begin{figure}
%   \includegraphics[width=\linewidth]{RE3.jpg}
%   \caption{GPT-FinRE pipeline flow}
%   \label{fig:Entity and Relation types in REFinD dataset}
% \end{figure}

\subsection{Prompt Construction}

We construct a prompt for each given test example,which is fed to the GPT models. Each prompt consists of the following components.

Task Description and Predefined Classes : We provide a succinct overview of the RE task description and the subset of predefined classes R, denoted by O. This subset is all possible relations exist between entity types of e1 and e2. The model is explicitly asked to output the relation, which belongs to the O. Otherwise, the model will output "no relation". 

K-shot Demonstrations : In the demonstration part, we reformulate each example by first showing the input prompt x-demo = Prompt(C, e1, e2) and the relation label y-demo. 

Test Input : Similar to the demonstrations, we offer the test input prompt x-test, and GPT models are expected to generate the corresponding relation y-test.

\subsection{Retrieval Systems}
We have employed two retrieval strategies to find top K relevant in-context learning demonstrations / examples from training data for a given test example.
\subsubsection{ KNN with OpenAI Embeddings}
Since ICL demonstrations closer to the test sample in the embedding space result in more consistent and robust performance \cite{DBLP:journals/corr/abs-2101-06804}. We utilized the KNN to retrieve the most similar examples in the training set as the few-shot demonstrations for each test example. As this learning-free dense retriever relies on the choice of the embedding space, we used OpenAI embeddings (text-embedding-ada-002) to obtain example representations. For similarity search, we used FAISS tool \cite{johnson2019billion}.
\subsubsection{ EPR (Efficient Prompt Retrieval)}
This learning-based dense retriever is trained to retrieve a better singleton in-context example \cite{rubin-etal-2022-learning} , and Top-K most similar examples are selected in the inference stage. This method for retrieving prompts for in-context learning uses annotated data and a LM. Given an input-output pair, it estimates the probability of the output given the input and a candidate training example as the prompt, and labels training examples as positive or negative based on this probability. It then trains an efficient dense retriever from this data, which is used to retrieve training examples as prompts at test time. Due to limited access to OpenAI, we have used the gpt-neo-2.7B model \cite{gao2020pile} as our choice of LM. 
  \begin{figure*}
  \includegraphics[width=10cm,height=6cm]{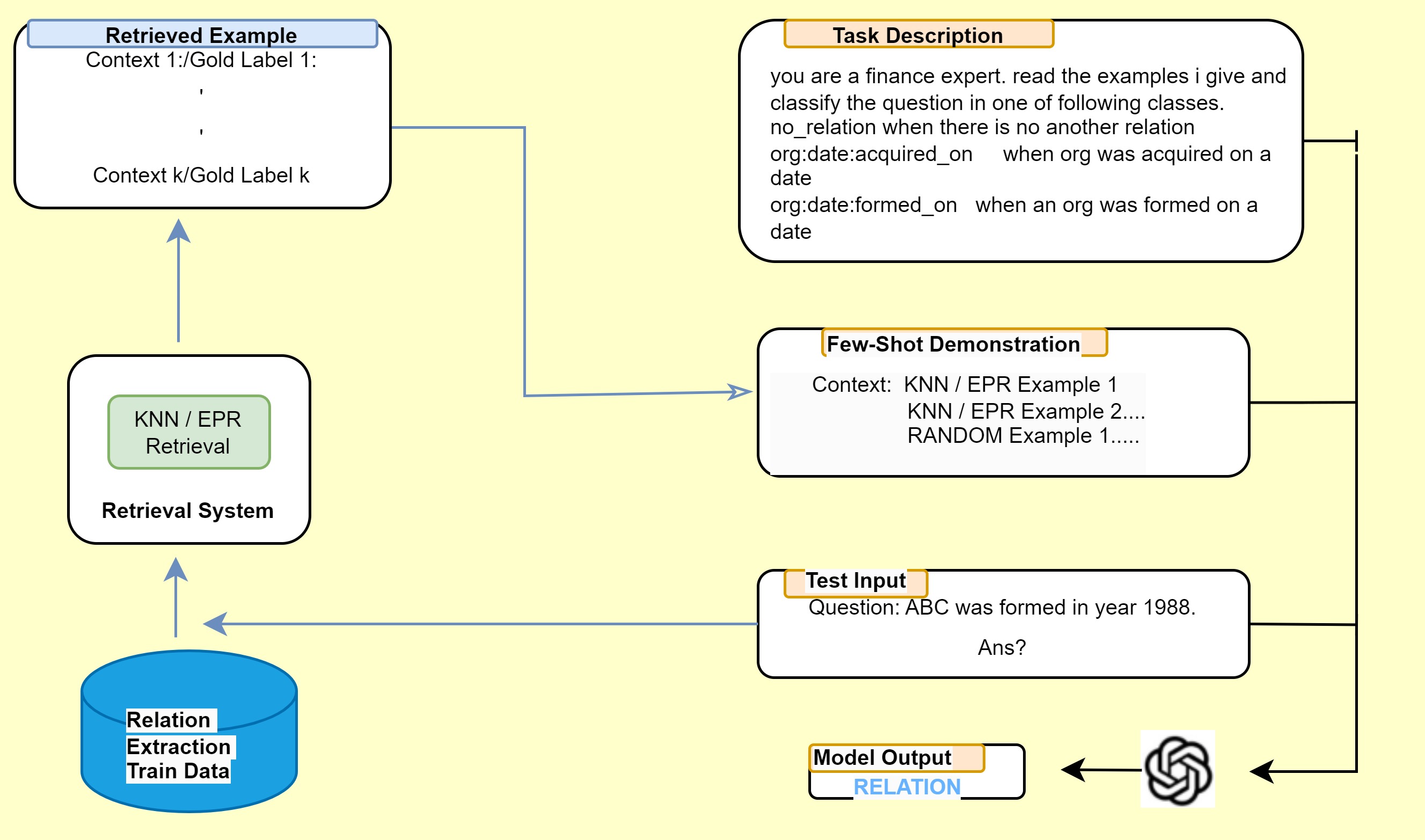}
  \caption{GPT-FinRE pipeline flow}
\end{figure*}
\subsubsection{Random Class Examples}
Along with KNN / EPR based examples, we also added K examples randomly for each possible class between two entity types to add more variety in the final prompt.

\section{Experiments}
Due to limited access and cost associated with OpenAI, We performed 4 primary experiments on the test dataset. We tried various rule based heuristics to improve the F1-score, but it didn't work as expected. We used retriever implementations from \footnote{\url{https://github.com/HKUNLP/icl-ceil}} . 

\section{Results}
The results are shown in Table-1. Our best F1-Score is 0.718. We got 4th position in the shared-task. We find that GPT 4 performs better than GPT 3.5 Turbo. We also find that learning based retriever (EPR) outperforms learning-free retriever (KNN with OpenAI embeddigs).

\section{Future Work}
In future we want to utilize GPT 4 for EPR. We also want to use different retrieval approaches such as Compositional Exemplars for In-context Learning (CEIL)\cite{ye2023ceil}.

\section{Conclusion}
This work explores the potential of GPT + ICL on Financial Relation Extraction (REFinD dataset). We used two retrieval mechanisms to find similar examples: (1) KNN with OpenAI Embeddings (2) EPR. We tried two different GPT models: (1) GPT 3.5 Turbo and GPT 4. The experimental results show that GPT 4 with learning based retriever EPR is giving the best F1-Score of 0.718.

\bibliographystyle{ACM-Reference-Format}
\bibliography{sample-base}

\end{document}